\documentclass[10pt,twocolumn,letterpaper]{article}
\pdfoutput=1
\usepackage{cvpr}
\usepackage{times}
\usepackage{epsfig}
\usepackage{graphicx}
\usepackage{amsmath}
\usepackage{amssymb}

\usepackage{url}
\usepackage{epstopdf}
\usepackage{wrapfig}
\usepackage{multirow}
\usepackage{caption}
\usepackage{changepage}

\usepackage[breaklinks=true,bookmarks=false]{hyperref}

\cvprfinalcopy 

\def\cvprPaperID{2106} 

\newcommand {\x}{\mathbf{x}}
\newcommand {\y}{\mathbf{y}}

\newcommand {\ofst}{\mathbf{u}}

\newcommand {\wgt}{\mathbf{w}}
\newcommand {\z}{z}
\newcommand {\vote}{\Omega}
\newcommand {\clust}{\mathcal{C}}

\ifcvprfinal\fi
\begin{document}

\title{Detecting Repeating Objects using Patch Correlation Analysis}

\author{Inbar Huberman  ~~~~~~~~~~ Raanan Fattal \\
School of Computer Science and Engineering\\
The Hebrew University of Jerusalem, Israel \\
}


\maketitle

\begin{abstract}
In this paper we describe a new method for detecting and counting a repeating object in an image. While the method relies on a fairly sophisticated deformable part model, unlike existing techniques it estimates the model parameters in an unsupervised fashion thus alleviating the need for a user-annotated training data and avoiding the associated specificity. This automatic fitting process is carried out by exploiting the recurrence of small image patches associated with the repeating object and analyzing their spatial correlation. The analysis allows us to reject outlier patches, recover the visual and shape parameters of the part model, and detect the object instances efficiently.

In order to achieve a practical system which is able to cope with diverse images, we describe a simple and intuitive active-learning procedure that updates the object classification by querying the user on very few carefully chosen marginal classifications. Evaluation of the new method against the state-of-the-art techniques demonstrates its ability to achieve higher accuracy through a better user experience.
\end{abstract}

\section{Introduction}

High object-count tasks are often encountered in industrial and scientific applications such as product inspection in manufacturing lines and cell counting for research and clinical purposes. These tasks typically require a considerable amount of repetitive human effort, and in many cases, a high degree of expertise. Automating the detection of objects using computerized vision is a highly challenging problem due to the visual complexity arising from irregular arrangement of the objects, variability in shape and illumination, mutual occlusions and similarity to other elements in the scene.

The object recognition literature in this context divides into two approaches. The first class of methods detect objects instances based on the raw response of various visual descriptors~\cite{dalal05,Gall09,Ke04,Nattkemper02} or on more detailed and generative object models~\cite{Mikolajczyk04,Felzenszwalb05,Wu08,Barinova10,Agarwal04,Leibe08}. These methods assume the objects are well-resolved and identifiable in the image. The second class of methods~\cite{Cho99,Chan08,Lempitsky10,Idrees13} is geared towards massively-populated images, e.g., human crowds, where individual objects consist of very few pixels. These methods typically regress the object density based on various texture descriptors, and estimate the count by integrating the density. Unlike the previous category, typically these methods do not localize individual objects. Both classes, however, involve training and typically require a large number of example images along with the locations or number of the objects they contain. The learnt models often show a high degree of specificity to the trained data, e.g, for a particular type of cell culture, and hence offer a limited use with more diversity scenarios.

In this paper we present a new method for localizing and counting one or more distinct objects in an image with no prior training stage. The method can operate on as little data as the input image itself, thus alleviating the need for large annotated training sets and offering a wider applicability compared to data-specific trained systems. The method relies on the deformable part-based model (DPM)~\cite{Fischler73,Felzenszwalb05} to detect the object of interest and enjoys the model's tolerance to moderate geometric deformations. While this model is rich in parameters and requires non-trivial training efforts, the key idea behind our approach is to exploit the sheer number of object appearances in the image to \emph{automatically} recover its parameters.


Similarly to other patch-based similarity measures~\cite{BoimanI06} we abstract the image content by considering small image patches as visual descriptors for the object parts and identify the ones associated with the repeating object by extracting highly-recurring windows in the image. Thus, we avoid the need to manually choose or pre-learn specific visual descriptors for the DPM. While the recurrent patches are likely to be part of the repeating object and can serve for its detection, some of the patches found may not be related to the object of interest and, at the same time, a single object occurrence is likely to stimulate the response of several patches. Thus, in order to disambiguate the occurrences and reject outliers we further search for a structural dependency between the patches, namely the springs of the DPM. Here again, we utilize the power hidden in the repetitiveness in the image which gives rise to meaningful auto- and cross-correlation functions of the patches occurrences that, in turn, allow us to derive spatial relations between the object parts. Besides requiring the user to provide the object scale, this procedure recovers the DPM parameters automatically and requires no training stage or data.

The recovered DPM provides a likelihood estimate for object occurrences across the image. We finally determine whether the object of interest appears in each potential location using a low-dimensional linear classifier which is tuned to the input image. The classifier's parameters are found by an intuitive active-learning procedure in which the user is asked to validate the classification of few carefully chosen instances. This user-assisted approach provides a practical tool for object counting and detection, without requiring any data beyond the input image, and is capable of adapting to the image particularities via a minimal user input. Evaluation against state-of-the-art shows that our object detection mechanism achieves higher accuracy on established datasets as well as our user assisted procedure fulfills this potential with very little input from the user.

\subsection{Related Work}

Object recognition is one of the most studied problems in computer vision. We focus here on works which are closely-related to object counting.

As noted above, one of the main paradigms for object detection consists of extracting various low- and mid-level visual descriptors from the image and using them to predict the existence of an object across the image. Example features used are: local histograms of oriented gradients~\cite{dalal05}, Hough transform~\cite{Gall09}, scale-invariant feature transform~\cite{Ke04}, and neural networks~\cite{Nattkemper02}. The predicting is done by some classifier, e.g., SVM, which is pre-trained on a database of images annotated with object locations. Models that rely on a more sophisticated description of the object are able to achieve higher classification accuracy and robustness to defomations and occlusions. For example, a hierarchy of part detectors are learned together with a pixel-level segmentor in~\cite{Wu08}. Higher-level geometric constraints are added to the Hough transform in~\cite{Barinova10}. Agrwal et al.~\cite{Agarwal04} learn a visual vocabulary of the object parts along with their spatial relations, whereas Leibe et al.~\cite{Leibe08} learn their individual spatial distribution. A full deformable part-based model is used for human detection in~\cite{Felzenszwalb05,Mikolajczyk04,Felzenszwalb10}. While these methods achieve notable results, they consist of parameter-rich models that require a non-trivial amount of training data and effort. In the context of texture segmentation, a fully-unsupervised generative tree model is described in~\cite{Ahuja07}.

A second line of works tackles the regime where the repeating object is under-resolved consisting of few pixels. These works provide count estimates (without their spatial locations) by calculating the object density in the image. The latter is typically regressed over various texture attributes and the objects number is obtained by integrating the density. Chan et al.~\cite{Chan08} feed edge and texture features into a Gaussian process which predicts the number of pedestrians in video sequences. An extension that collects texture and periodicity descriptors from multiple scales is described in~\cite{Idrees13}. While targeting less densely populated images, Cho et al.~\cite{Cho99} estimate crowd density using a neural network stimulated by various edge detectors. Training these methods requires set of images along with the number of objects they contain. Hence each example provides a single, or very few, constraints over the regression parameters. Lempitsky and Zisserman~\cite{Lempitsky10} also assume the objects are well resolved and learn the density maps from user-provided set of locations, thus reducing the number of training images needed. This approach is further accelerated using regression trees in~\cite{Fiaschi12}.

To tackle more practical scenarios and avoid the need for training data and its associated bias, Arteta et al.~\cite{Arteta14} describe an interactive procedure in which a linear regression is used over dimensionally-reduced image descriptors. Yao et al.~\cite{Yao12} describe an interactive video annotation method based on Hough trees and provides live feedback to minimize human effort. The CellC~\cite{CellC} and ImageJ~\cite{ImageJ} are two open-source interactive GUIs designed for segmenting and counting cells in fluorescence microscope images. These methods use intensity-based thresholds defined by the user. Sommer et al.~\cite{Sommer11} extracted sophisticated texture descriptors in where random forests are used for classification.

Similarly to our work Torii et al.~\cite{Torii13} exploit repetitiveness in images, however, their work focuses on place recognition task and their algorithm estimates the abundance of various visual words, regardless of their relation to any particular shape or object in the scene. Leung and Malik~\cite{Thomas6} extract segments consisting of a repeating patten under affine transformations.


\section{New Method}
\label{sec:method}

The main idea behind the new method is to avoid the costs involved in the conventional training used for object detection by exploiting the fact that the input image contains multiple appearances of the object of interest. The method derived in this section carries out this idea using a fairly-sophisticated visual model, namely the deformable part model in two steps. In the first step, given the user specified bounding box of the object, the DPM components are automatically recovered by extracting recurrent patches in the image and analyzing their spatial correlation. In the second step, the existence of the object at the potential locations indicated by the DPM is determined using a linear classifier. The latter is adapted to the input data by an intuitive active-learning procedure which requires the user to provide feedback on a very few marginal instances.

In the next two sections we describe the procedure for recovering the DPM visual descriptors (nodes) and their spatial arrangement (springs). In Section~\ref{sec:detect} we explain how the objects in the image are detected using the DPM found.

\begin{figure}[t]
\centering
\includegraphics[width=3.4in]{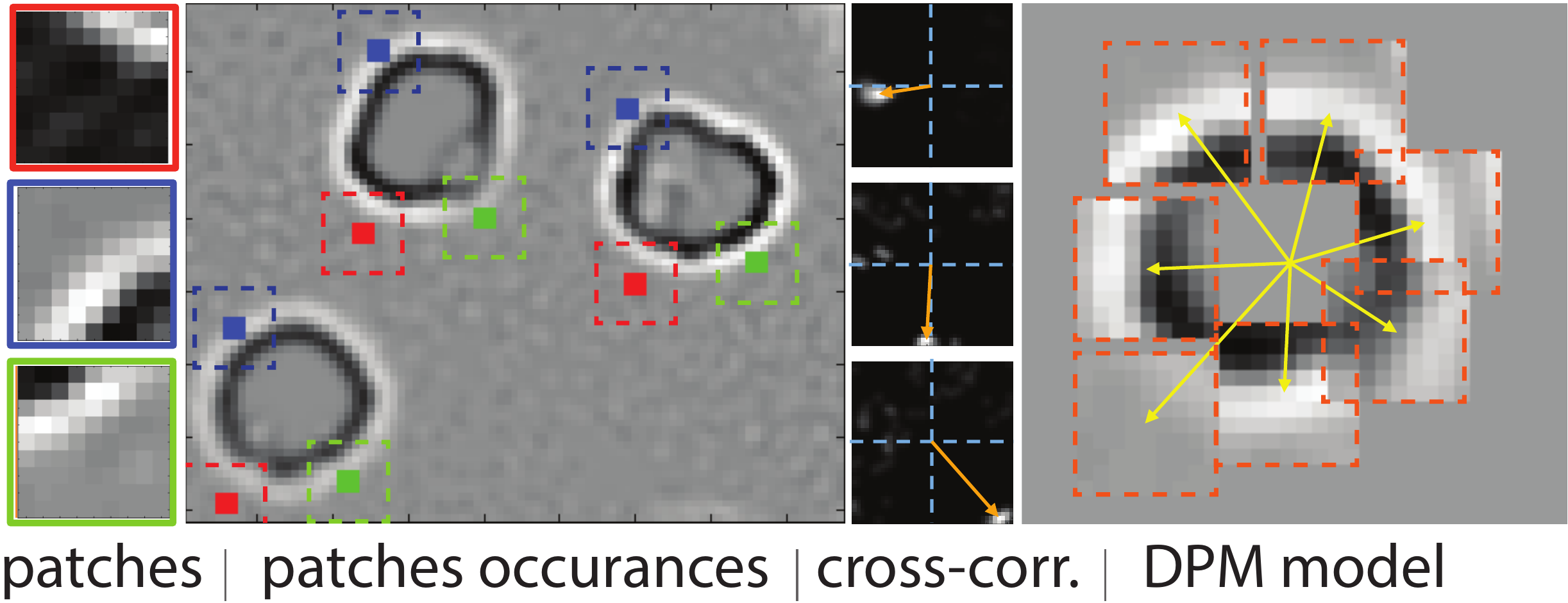}
\caption{Left to right are: three recurrent patches and their occurrences in the image. The spatial offset between the occurrences indicated by the peak's location (orange arrow) in the cross-correlation functions. A visualization of the DPM recovered by our method.}
\label{fig:model}
\end{figure}

\subsection{Extracting Recurrent Patches}
\label{sec:search}


Given an input image $I(\x)$ containing multiple instances of an object, we expect to find a number of recurring patches that take part in the object's pattern.
We extract these patches from the image and use them as the visual descriptors of the object's DPM. The patches are extracted by a simple iterative procedure in which we consider, at each step, multiple random patches of the image, and pick the one with the maximal number of appearances. In this section we explain this procedure in detail. As the first step we consider a small square window centered around a random coordinate as a candidate patch $p$, and compute its cross-correlation\footnote{cross-correlation value between two image patches $p$ and $q$ is defined by $\langle p-\mu_p,q-\mu_q\rangle / (\sigma_p \sigma_q)$ where $\langle\cdot,\cdot\rangle$ denotes the dot-product operator, $\mu$ denotes the patch average and $\sigma$ its standard-deviation. The auto-correlation of $p$ is obtained by setting $q=p$.} function, $\rho(\x)$, with every patch in the image (centered around $\x$).

The \emph{occurrence map} $\z(\x)$ of $p$ is defined by a maximum suppression over $\rho(\x)$, specifically; we set $\z(\x)\!=\!1$ where $\rho(\x) \!>\! 1\!-\!\varepsilon$, unless $\rho(\x)>\rho(\y)$ at any pixel $\y$ inside a window around $\x$ of the same dimensions as the patches used, and $\z(\x)\!=\!0$ elsewhere. We use $\varepsilon = 1/20$ in all the tests reported but expect this value to increase in presence of high noise levels. The patch sizes we use is 9-by-9 pixels assuming the object is three times this size. Therefore, similarly to~\cite{Arteta14}, we ask the user to bound one of the objects in the image with a box and scale the image accordingly.

Recall that the occurrence map $\z(\x)$ is a binary map, where $z(x) = 1$ indicates the pixels in which the image $I(\x)$ and the patch $p$ show a strong structural resemblance and is zero otherwise. Thus, we measure $p$'s frequency in the image by $\sum_\x\z(\x)$. We repeat this step several times (consider 30 candidate patches in our implementation) and add the patch with the highest frequency to a list of \emph{recurrent-patches}, $p_1,p_2,...$ and keep the corresponding occurrence maps $\z_1,\z_2,...$. This procedure is repeated until the frequency of the patch found falls below some fraction of the maximal frequency encountered in the previous steps ($30\%$ in our implementation). To avoid multiple selections of the same or very similar patches, we exclude pixels that neighbor occurrences of previously selected patches. More formally, at the $j$-th step we skip patches that overlap pixels $\x$ in which $\sum_{i=1}^{j-1}\z_i(\x) > 0$. We further accelerate the process by skipping candidate patches with low variance due to their inability to describe shape. Figure~\ref{fig:model} shows three patches recovered in this process and their occurrence maps.

While not dealing with object counting, Shechtman and Irani~\cite{Shechtman07} identify large shapes based on similarity between the patches that compose it. Our work relies on a different observations, where the patches are expected to appear once in every object occurrence.


\begin{figure}[t]
\centering
\includegraphics[height=1.27in]{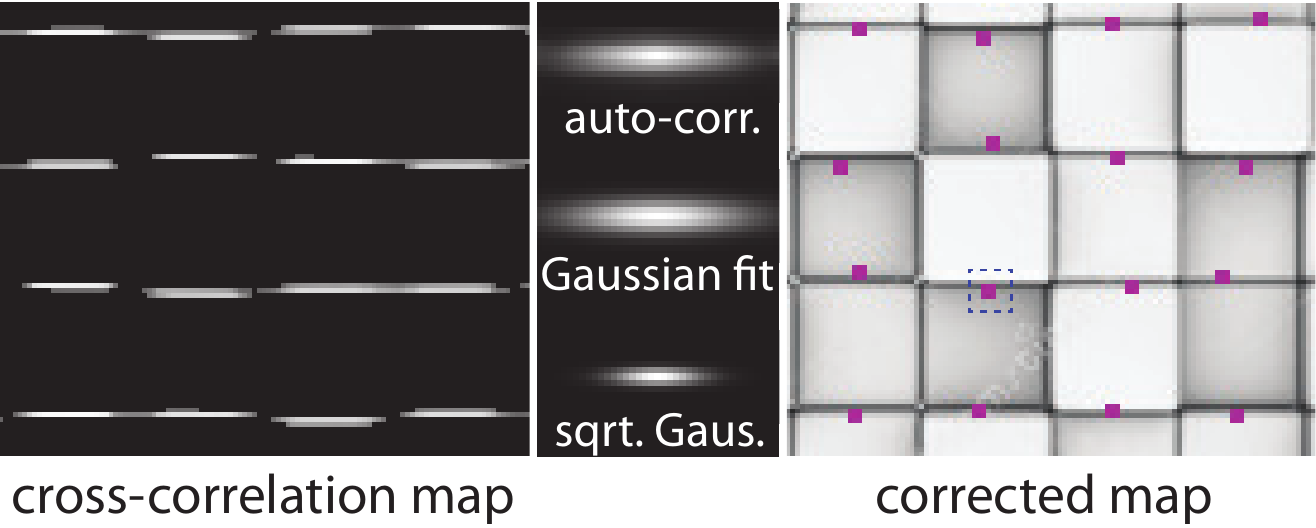}
\caption{Left: correlation map of a patch containing a straight edge (blue dashed line). Center: the auto-correlation function of this map. Beneath is its Gaussian approximation and at the button is the square-root of the Gaussian approximation. Right: corrected occurrence map (purple).}
\label{fig:edge}
\end{figure}

\subsection{Structure from Patch Correlation Analysis}
\label{sec:corr}

It is highly likely that many of the recurrent patches found correspond to different parts of the repeating object and hence appear in a consistent spatial arrangement, see Figure~\ref{fig:model} as example. We detect this spatial dependency using a patch correlation analysis and use it to define the spring parameters of the DPM. This step allows us to associate patch occurrences with the occurrence of a unique object as well as reject outlier patches found in the previous step.

Here again we exploit the object's repetitiveness in the image which provides meaningful auto- and cross-correlation functions $\tau_{ij}$ between pairs of occurrence maps $\z_i,\z_j$, where $1 \leq i,j \leq n$ and $n$ is the number of patches extracted. In principle, if the $i$-th and $j$-th patches correspond to two parts of the repeating object, both of them are expected to respond to the same occurrences of the object in the image. Each of these responses will be recorded both in $\z_i$ and $\z_j$ but at a different spatial offset which corresponds to the patches offsets in the object. This spatial dependency in the occurrence maps will trigger a peak in their cross-correlation $\tau_{ij}(\x)$ around the coordinate $\x$ that is equal to the spatial offset vector. Note that while false responses may exist in the occurrence maps, it is less likely that these occurrences will exhibit a strong mutual correlation as to corrupt $\tau_{ij}(\x)$.

We detect strongly correlated patches by measuring the ratio between the maximal value in $\tau_{ij}(\x)$ and the second-largest value which acts as a local optimum (greater from its eight nearest points). If this ratio is above some factor (two in our implementation) we mark the patch pair $(i,j)$ as being correlated and extract their characteristic spatial offset by $\ofst_{ij} = \text{argmax}_{\x} \tau_{ij}(\x)$. Figure~\ref{fig:model} shows patch offsets recovered from the correlation functions. At the end of this procedure each patch has between zero to $n-1$ paired patches. Finally, we discard patches with less than $n/10$ pairs.

As we noted earlier, the image may contain recurrent patches that do not belong to the object of interest. Typically, such patches result from straight edges which are abundant in natural images or the existence of an additional repeating pattern in the image. In the first case (edges) or repeating objects of the patch size (noise), the patches are not expected to show a strong correlation with other patches, hence be discarded due to their insufficient number of patch pairs. In Section~\ref{sec:detect} we explain how we cope with images that contain more than one large repeating object.

A straight edge \emph{in the object} of interest poses a different problem where the relative location of the patches constructing it is not well defined (along the edge). In practice, as shown in Figure~\ref{fig:edge}, the correlation map, $\rho_i(\x)$ of such patches contains linear features instead of well-defined peaks. In order to come up with a valid occurrence map for such patches, we identify the locations of each linear feature in $\rho_i(\x)$ using template matching with a single instance, $r(\x)$, of these features. We recover $r(\x)$ by first computing the auto-correlation function of $\rho_i(\x)$, namely $R(\x)$, which contains a single instance convolved with itself, i.e., $r(\x)*r(\x)$ (in case of a symmetric function). By approximating the function $R(\x)$ with a 2D Gaussian $G_\sigma(\x)$ using a PCA, we approximate $r(\x)$ using $G_{\sigma/2}(\x)$ which is the square-root of $G_\sigma(\x)$ with respect to the convolution operation. Finally, we apply a maximum suppression over $\rho_i(\x)* r(\x) $ to obtain the new occurrence map $\z_i(\x)$. We compute the auto-correlation function of every patch found, and apply this procedure only in cases where the Gaussian approximation exhibits high eccentricity (above 2). Figure~\ref{fig:edge} shows the results of this correction step.

\bf{Planner Embedding. }\normalfont The set of recurring patches found so far correspond to vertices of an incomplete graph whose edges are the strongly-correlated patch pairs. Every edge in the graph is associated with the average spatial offset vector $\ofst_{ij}$ between the patches. As the final step of constructing the shape model, we convert these relative relations into a consistent set of coordinates in the plane which we use below for efficient detection of potential object occurrences.

The planner embedding problem we are faced with can be solved by a straightforward application of the \emph{locally-linear embedding} (LLE) technique~\cite{Roweis00} in which we seek for global vertex coordinates, $\x_1,...,\x_n$, that agree, as much as possible, with the spatial offsets $\ofst_{ij}$ by minimizing
\begin{equation}
\min_{\x_1,...,\x_n} \sum_{(i,j) \in E} \| \x_i\!-\!\x_j\!-\!\ofst_{ij} \|^2,
\end{equation}
where $E$ denotes the set of edges in the graph (pairs of correlated patches). Upon differentiation w.r.t, $\x_1,...,\x_n$, we obtain a linear system over the $x$ and $y$ coordinates of the vertices. Figure~\ref{fig:model} depicts the planner arrangement found in this step. Note, that while the object instances can appear under different deformations, the highly-correlated patches we are considering correspond to adjacent object parts that experience only a small amount of the deformation. The resulting coordinates correspond to consistent biases in the location of every patch in the object.

Note that our DPM is not restricted to a single version of an object or even a single object (see Figure~\ref{fig:features_1}). The DPM's graph can contain alternative nodes for certain parts of the object, as well as multiple connected components representing different uncorrelated objects.

\begin{figure}[t]
\centering
\includegraphics[width=3.4in]{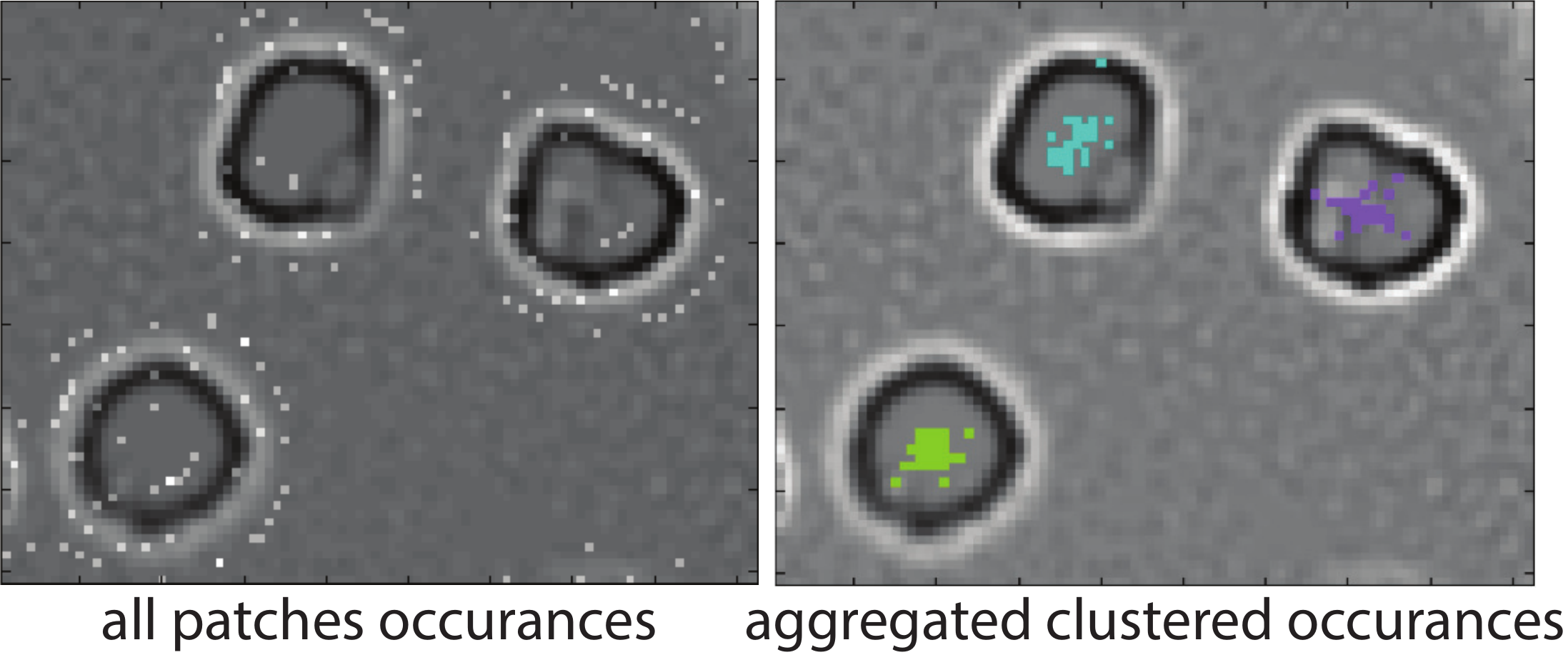}
\caption{All the patch occurrences found and their aggregation according to their embedded coordinates.}
\label{fig:occurrences}
\end{figure}

\subsection{Object Detection}
\label{sec:detect}

In principle every response in every occurrence map computed provides an evidence for the appearance of an object. We use the recovered DPM to identify locations in the image in which there is a consensus among one or more such evidences. We further collect a small number of basic measures from each of these possible appearances and feed this data, as a feature vector, into a linear classifier which makes the final judgment whether an object appears in each of the suspected locations. In order to adapt the algorithm to the particularities of the input image and object shape, we allow the user to fine-tune the classifier via a small number of carefully-selected queries.

\bf{Potential Occurrences. } \normalfont In case a DPM has a tree structure the detection of object instances can be performed using dynamic programming~\cite{Felzenszwalb05} in $O(nN)$ operations, where $N$ is the number of image pixels (and $n$ is the number of parts/patches). As noted above, we utilize the patch occurrence maps computed earlier to accelerate the search down to $O(nm)$ where $m$ is the number of object occurrences. This is done by collecting the coordinates in which the occurrence maps responded and shifting them to the object center as defined by the embedded coordinates, namely $ \vote_i\!=\!\{\y\!:\!\z_i(\y\!-\!\x_i)\!=\!1\}$. While each instance can appear with its own deformation, this transformation eliminates the consistent bias in the patches location in the object and, as shown in Figure~\ref{fig:occurrences}, leads to substantial aggregation of the responses. Finally, in order to extract the consensus locations, we apply a multi-model RANSAC clustering~\cite{Kenney05} over $ \vote=\cup_{i=1}^n \vote_i $ with $\sigma = 20$ (two thirds of the object size). At every cluster $k$ we store its center coordinate $c_k$, the list of patches assigned to it $\mathcal{C}^k$, and their coordinates $\y_j$, where $j \in \mathcal{C}^k$.

\begin{figure}[t]
\centering
\includegraphics[width=2.8in]{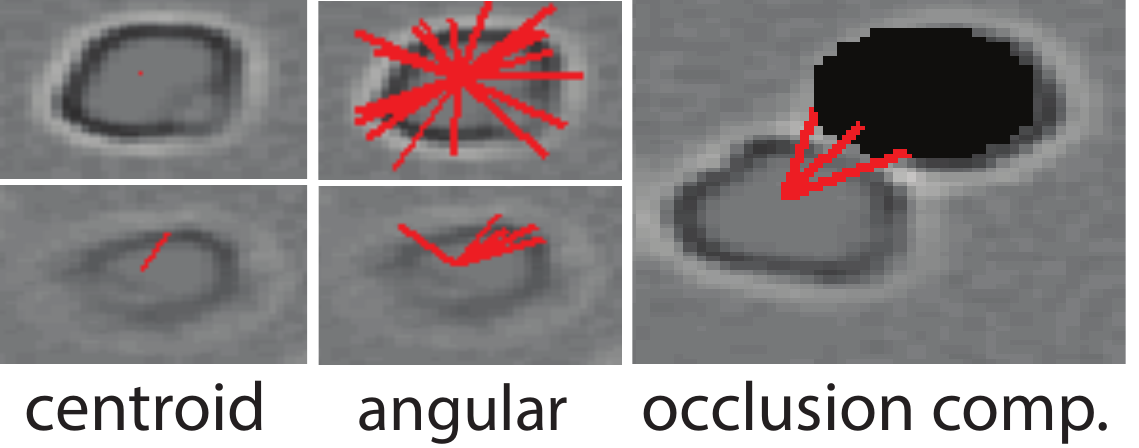}
\caption{Features used for classification (left-to-right): difference between patches centroid and cluster center, angular bins occupancy and rays reaching to occluder pixels in the feature maps.}

\label{fig:features}
\end{figure}

\bf{Occurrence Descriptors. }\normalfont In order to determine whether the clusters found indicate a true object appearance or not while coping with the unique characteristics of the input image, we resort to a fairly low-dimensional linear separator to perform this classification. The classification is based on the following basic fidelity measures extracted at every cluster $k$ found: (i) number of patches in the cluster, (ii) average correlation values between patches and the image (given by $\rho$), (iii) the average deformation in the cluster, given by
\begin{equation}\vspace{-0.05in}
\sum_{j,j'\in \clust^k, (i_j,i_{j'}) \in E} \| \y_j\!-\!\y_{j'}\!-\!\ofst_{i_j,i_{j'}} \| / \big|\{j,j'\in \clust^k, (i_j,i_j') \in E\}\big|,
\end{equation}
(iv) the average distance between the centroid of the patches present in a cluster and its center $\| \sum_{j\in\clust^k} \y_j/ |\clust^k| - c_k\|$, which is expected to vanish when all patches participate.


In order to compensate clusters near the image border we add (v) the distance between the cluster center, $c_k$, and the closest boundary (as long as the distance falls below the object radius).



\begin{figure}[t]
\centering
\includegraphics[width=3.3in]{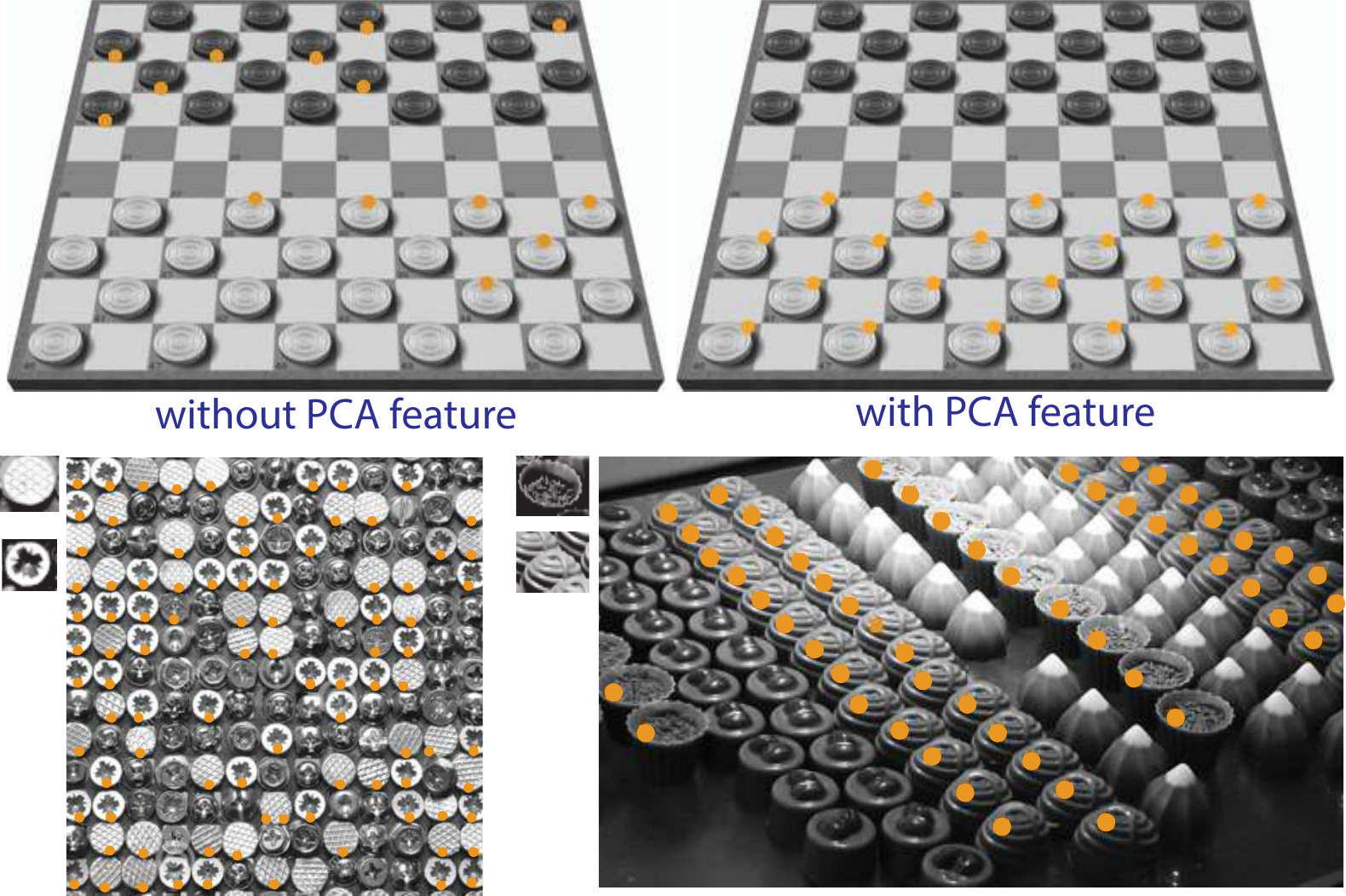}
\caption{PCA-based patch composition demonstration: PCA descriptors allow us to differentiate between distinct types of repetitive objects. The top example is the result obtained with and without these descriptors when insisting to count only the white pieces. In the bottom examples we count two specific types of screws and pralines and exclude the rest (with PCA descriptors used).}
\label{fig:features_1}
\end{figure}

Repeating patterns in the image, besides the object of interest, are likely to consist of a different composition of patches. Hence, in order to differentiate between different populations of objects we apply PCA over the set of vectors that indicate which patch participates in each cluster, $v^k_i=1$ if $i \in \clust^k$ and zero otherwise. Thus, as the (vii)-(ix) features, we compute the $\langle v^k, a_1 \rangle, \langle v^k, a_2 \rangle$ and $\langle v^k, a_3 \rangle$, where $a_1, a_2$ and $a_3$ are the three most active principal directions. Figure~\ref{fig:features} visualizes some of the features we described and Figure~\ref{fig:features_1} demonstrates the improved ability to discriminate between distinct types of objects based on the last feature.

Occluded objects are another case requiring compensation due to missing patches. If the occluding object is an instance of the object being counted, its presence is likely to be captured by one of the clusters found. However, since none of the objects are identified at this stage, we only look at the feature values at locations of potential occluders. More specificity, we generate \emph{feature maps} containing the feature values extracted from each cluster rendered at pixels that the object is expected to occupy, i.e., circles located around the cluster centers $c_k$. We use the inputted object size as the diameter of the circles. The rest of the maps values are set to zero. Then, for each empty angular bin in a cluster we evaluate feature values from the maps and concatenate it to the cluster's feature vector describe above. Thus, the feature vectors we assign to each cluster, $f_k$, is a 18-dimensional vector. The values arriving from different bins are added together. As shown in Figure~\ref{fig:features} the sampled points are the endpoints of rays emanating from the cluster centers, passing through the empty angular bins, and extending to a distance equals to the object size.

\bf{Occurrence Classification. }\normalfont Finally, we determine whether each cluster corresponds to an object appearance or not by feeding its feature vector to a linear SVM, namely, $\text{sign}(\langle f_j,\wgt\rangle - b)$, where $\wgt$ and $b$ are the separation vector and offset value. As noted above, we allow the user to tune this decision making to the particularities of the input image by updating these parameters through the following query-based procedure. At the first step, we initiate $\wgt = (1,0,0,...)$, i.e., consider only the number of patches, and compute biases $b_{\min}$ and $b_{\max}$ that lead to all positive and all negative decisions. We extract 20 clusters that uniformly sample this range with their scores, $\langle f_j,\wgt\rangle$, and allow the user to change their classification by picking $b \in [b_{\min},b_{\max}]$ using a slider. The classifications are conveyed by rendering a red or green frame around each of the 20 objects.

Given the current $\wgt$ and $b$ we set $\delta^\pm\!=\!\min \{ |b\!-\!b_{\min}|,|b\!-\!b_{\max}| \}$ and proceed with the following procedure. We randomly sample seven clusters that score $ b < \langle f_k,\wgt \rangle < b\!+\!\delta^+/2$ and another three with $  b\!+\!\delta^+/2 < \langle f_k,\wgt \rangle < b\!+\!\delta^+$. We do the same for the negative case; sample seven and three clusters whose score falls inside $ [b\!-\!\delta^-/2,b]$ and $ [b\!-\!\delta^-,b\!-\!\delta^-/2]$. The classification of all the 20 clusters are presented to the user who can change them by clicking inside the colored frames that mark their classification. While the user does not have to click all 20 clusters, the proper classification of all these clusters is inputted in this step. We use this user input as well as all the clusters scoring above $ b\!+\!\delta^+$ and below $ b\!-\!\delta^-$ as tagged points in a soft-max SVM to obtain new $\wgt$ and $b$. The classification of clusters which are close to the separation margin is less reliable and therefore we devote more user validation
over these clusters by biasing the cluster selection towards the margin (14 clusters out of 20).

\begin{table}[t]
\centering
\begin{tabular}{ lccc }
  & dataset & avr. err. & std err. \\
\hline
\hline
\multirow{3}{4em}{Lempitsky Zisserman} & small & 2.9/3.2 & 2.85/2.71  \\
																&	large  & 12.9/11.6  & 4.4/5.7 \\
																&	lg.-sm.  & 3.65/4.91    & 4.4/4.17 \\
\hline
\multirow{2}{3em}{Hough} & small & 5.19 & 3.05 \\
									&	large  & 8.56    & 6.69 \\
\hline
\multirow{2}{3em}{Template Matching} & small & 12.25 & 9.4 \\
									&	large  & 13.75    & 9.8 \\
\hline
\multirow{3}{4em}{Splitted Template Matching} & small & 12.5 & 13.3 \\
									&	large  & 9.8    & 10.4 \\
                                    &	lg.-sm.  & 10    & 7.2 \\
\hline
\multirow{3}{3em}{our} & small & \bf{1.5} & \bf{1.2} \\
									&	large  & \bf{4.25}     & \bf{2.5} \\
									&	lg.-sm.  & \bf{1.93}    & \bf{1.6} \\
\end{tabular}
  \captionof{table}{Small refers to the cell images produced by~\cite{Lehmussola07} and large to the addition of few false larger cells into the images in this dataset. The dataset titled lg.-sm. refers to training on the mixed cells (including the large cells) and testing on the images containing only small cells. The two error values in Lempitsky and Zisserman's column correspond to the two metrics they use. Each image in the datasets contains 250 cells.}
  \label{tab:supervised}
\end{table}

We also decrease $\delta^+$ by a factor of two if the classification of the three extreme clusters in $[b\!+\!\delta^+/2,b\!+\!\delta^+]$ was not changed by the user, and increase $\delta^+$ by that factor if it was. The same goes for $\delta^-$ and $ [b\!-\!\delta^-,b\!-\!\delta^-/2]$. We repeat this  procedure with the updated values of $\wgt,b$ and $\delta^\pm$ until there are no corrections entered or not enough clusters can be found within $[b\!-\!\delta^-,b\!+\!\delta^+]$. By increasing and decreasing the interval $[b\!-\!\delta^-,b\!+\!\delta^+]$ we adapt online the criterion of marginal classifications and allow the system to perform both large and subtle updates in the classifier parameters.

In contrast to the method of Arteta et al~\cite{Arteta14}, this procedure does not require the user to inspect all the objects in the image and compare them to another image showing the system decisions. The user is also not required to outline regions or pinpoint items in the image.

\section{Results}
While the goal of this work is to derive a practical user-assisted counting technique, we start with a test assessing the algorithm's accuracy limit. In this test, given the object scale, we recover the DPM automatically, as described in Section~\ref{sec:method}, but replace the user-assisted procedure with a training stage in which we optimize the linear classifier over 16 simulated fluorescence microscope cells images taken from~\cite{Lehmussola07}, which were previously used to benchmark object counting algorithms~\cite{Lempitsky10,Arteta14}.

\begin{table}[t] \addtolength{\tabcolsep}{-3pt}
\begin{tabular}{ cccccccccc }
\#patches & corr. & deform. & cent. & ang. & PCA & borders &  occl.\\
\hline
\hline
77\% & 425\%  & 125\% & 61\% & 116\% & 138\% & 183\% & 135\%\\
\hline
\end{tabular}
\caption{Percentage of error increase when omitting each descriptor from the SVM classifier and testing on the lg.-sm. dataset.}
\label{tab:ignore}
\end{table}

Table~\ref{tab:supervised} summarizes the results and shows that our object detection mechanism has potential to achieve a higher accuracy compared to the supervised density estimation method of Lempitsky and Zisserman~\cite{Lempitsky10}.

Given the user supplied bounding box of the object, from which we derive the object scale, we compared our method with template matching methods, namely the classic approach~\cite{Briechle01} that uses a single template, and a newer one by Boiman and Irani~\cite{Boiman07} which breaks the template into smaller windows. As shown in Table~\ref{tab:supervised}, the DPM we extract from the image is more
accurate than these approaches as it uses patches whose recurrence and mutual correlation were recovered from the entire data and not deduced from a single instance. A single template may also contain irrelevant background features if not carefully selected.

\begin{table}[t]\addtolength{\tabcolsep}{-4pt}

\begin{adjustwidth}{-0.5cm}{}
\begin{tabular}{ccccccccccc}
\setlength{\tabcolsep}{-4pt}
 &  image & 1 & 2 & 3 & 4 &  5 & H1 & H0 & clicks & time \\
\hline
\hline
\multirow{5}{2em}{Arteta\\ et al.}
&  water & 31.3  & 20.9 & 29.8 & 20.7 & 18.7 & 4.8 & 28.2 & 16.2 & 1:46  \\
&  beer &  68.9  & 39.9 & 34.8 & 32.8 & 26.4 & 15.6 & 42.1 & 14.5 & 2:27  \\
&  pills & 26.1  & 20.6 & 11.4 & 4.3 & \bf{3} & 2.4 & \bf{3.2} & 16.1 & 2:08  \\
&  small & 23.8  & 21.7 & 15.6 & 18.6 & 13.9 & 1.7 & 37 & 24 & 2:48  \\
& large & 29.4  & 51.3 & 47.7 & 36.9 & 27.7 & 3.4 & 49.8 & 17.8 & 2:35  \\
\hline
\multirow{3}{2em}{CellC}
& water & 836 & 442 & 442 & 442 & 442 &   &   & 5 & \bf{0:25}  \\
& beer & 301  & 63 & 16 & 16 & 16 &  &   & 5 & \bf{0:27}  \\
& small & 19.3  & 9.7 & 9.7 & 8.9 & 29.9 &   &   & 5 & \bf{0:34}  \\
\hline
\multirow{5}{2em}{our}
& water &  \bf{3.5}  & \bf{3.1} & \bf{2.5} & \bf{2.5} & \bf{2.5} & \bf{0.3} & \bf{2.3} & \bf{3.1} & \bf{1:05} (0:17)  \\
& beer &  \bf{4.5}  & \bf{2.7} & \bf{2.8} & \bf{2.7} & \bf{2.7} & \bf{0.7} & \bf{2.6} & \bf{6.3} & \bf{2:08}  (0:21)\\
& pills &  \bf{9.8}  & \bf{5.7} & \bf{5.2} & \bf{4.1} & 3.25 & \bf{0.4} & 4.2 & \bf{14.1} & \bf{1:47} (0:08) \\
& small &  \bf{5.9}  & \bf{6.2} & \bf{4.4} & \bf{3.2} & \bf{0.4} & \bf{0.4} & \bf{2.2} & \bf{5.9} & \bf{2:18} (0:30) \\
& large &  \bf{4.7}  & \bf{5.8} & \bf{4.7} & \bf{3.8} & \bf{3.2} & \bf{0.6} & \bf{2.2} & \bf{7.3} & \bf{2:07} (0:25)\\

\end{tabular}
\captionof{table}{Counting errors at each iteration are shown in the first five rows. The final false positive H1 and negative H0 and the total number of mouse clicks are provided in the following rows. The total session time is given in the last row and the pre-process portion of our method is given in brackets (this time is included in the time row). All the images can be found in the supplementary material.}

\label{tab:userassist}
\end{adjustwidth}
\end{table}

Table~\ref{tab:supervised} also shows the results obtained by running a circular Hough transform. The relatively high error produced is a result of the fact that incomplete objects create a wide range of scores that make it hard to find a suitable threshold for detecting occurrences.


It is worth noting that when training both our and Lempitsky and Zisserman's systems on a dataset containing a few extra false large cells, the performance on images consisting of only small cells decreases. This specificity to the training data is avoided in the user-assisted approaches.

In order to verify that each descriptor in the feature vector contributes to the classification accuracy, we repeated the fully-supervised experiment eight more times. In each experiment we ignored one of the feature elements and calculated the average error. Table~\ref{tab:ignore} shows the increase in error compared to full feature vector.

In order to evaluate the user-assisted procedure against Arteta et al.'s we conducted a user study consisting of 15 participants and 5 images. Before asking the participants to use these systems, we demonstrated how each system operates by running an example session. The order of the methods and images, tested in this study, were chosen randomly for each participant. Table~\ref{tab:userassist} shows the average counting error at each iteration of these user-assisted procedure. The results show that our counting procedure capable of achieving more accurate counts at slightly shorter user time and effort (mouse clicks). Moreover, the average error on the cell images (small) appears to be very close to the optimal solution obtained by a fully-supervised training stage (see Table~\ref{tab:supervised}). We find the latter to be a positive indication that our update procedure is capable of reaching close-to-ideal separators. The results obtained from the CellC software~\cite{CellC} are also given in Table~\ref{tab:userassist} and appear to achieve a lower accuracy.

False positive and false negative errors cancel out when considering only the total object count, however these values are important when assessing the object localization error. Table~\ref{tab:userassist} indicates that our method manages to achieve a considerably higher accuracy in this respect. We further asked each of the participants, at the end of their session, which system they found more easy to work with and all, but one, preferred our query-based approach.

\begin{figure}[t]
  \centering
  \includegraphics[width=3in]{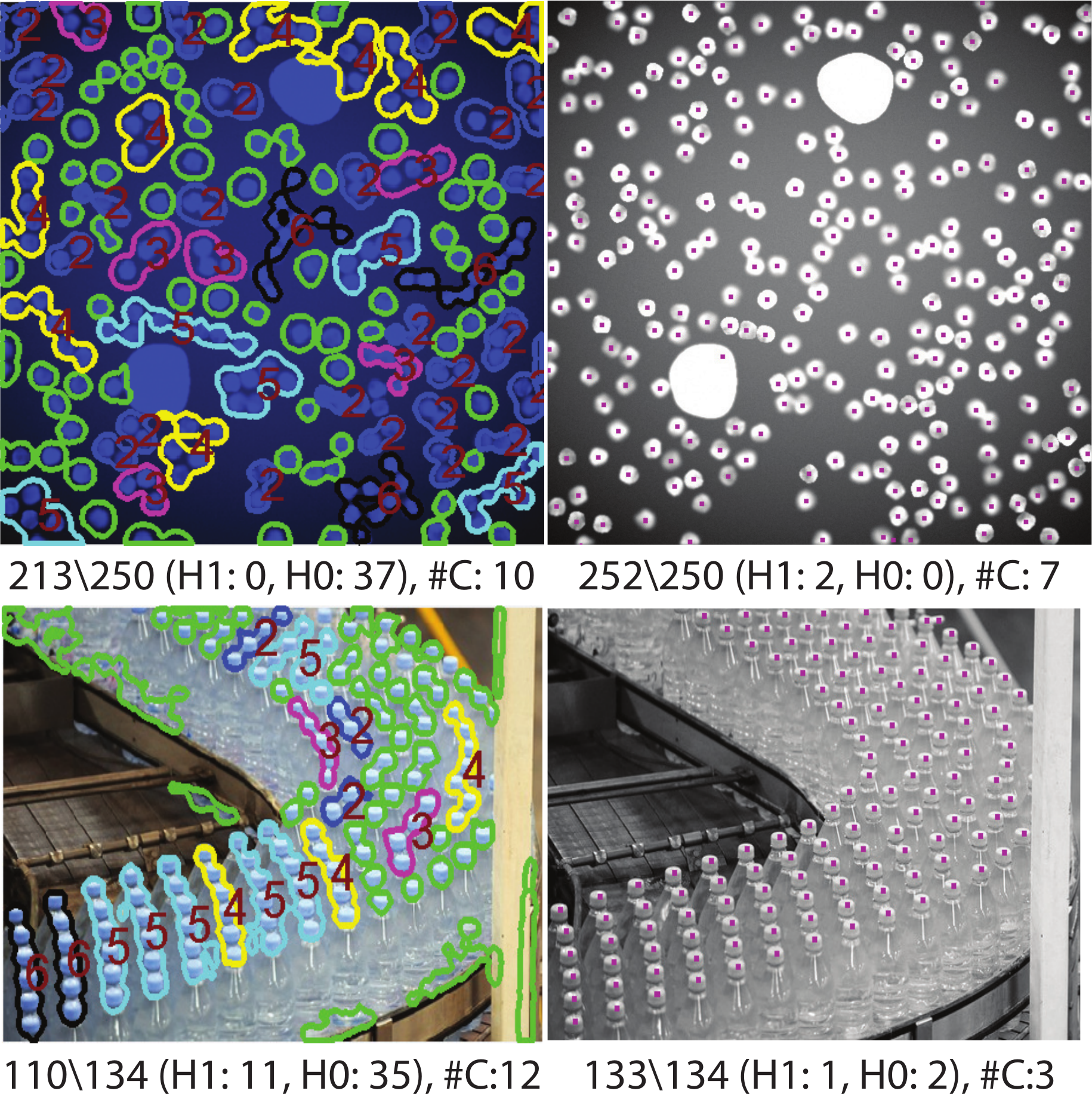}
  \captionof{figure}{Images produced by the method of Arteta et al. and our. In both cases less mouse clicks (indicated by `\#C') where used to produce our results.}
  \label{fig:examples}
  \vspace{0.15in}
\end{figure}

Finally, Figure~\ref{fig:examples} shows example images and the objects detected by the two methods. The water bottles image is an example where the object size changes due to perspective deformation. While the features we use namely, image patches, are not invariant to such transformation a sufficient number of instances at each size enables us to find enough identifying patches. Nonetheless, as we discuss below, we find this lack of invariance as well as rotational to be a drawback of our method.

In the supplementary material we provide all the results presented here as well as additional results, including  videos capturing a user interaction session.


%

\section{Conclusions}

We presented a new method for localizing and counting repeating objects in an image. While the new method consists of a fairly detailed DPM, it requires no training phase or data. They key idea that allow posing the model is to exploit the shear number of object repetitions in the image. Rather than resorting to content-specific feature spaces, we abstract the image content and use its own patches to search for recurrences. The DPM's shape is obtained by a patch correlation analysis which also serves us to efficiently identify potential object occurrences. Finally, we described a practical system which is capable of adapting to the unique characteristics of the image and the objects it contains. This approach consists of a simple and intuitive active-learning procedure which allows the user to correct the decision of the classifier over a small number of carefully-selected queries and without the need to inspect the entire input.
Results show that the new algorithm is capable of achieving more accurate estimates on images from a known benchmark dataset, and that this accuracy is more easily achieved using the new user interaction system.


Despite the advantages presented in the previous section, there are many scenarios in which our method is not expect to perform well. Clearly, images with severe occlusions or considerable object variation will prevent it from identifying a sufficient number of repeating patches and their occurrences, which will undermine our ability to detect meaningful spatial dependency. Experiments show that at least 20 object repetitions are required for the method to perform. As noted earlier, we do not cancel rotational or scaling transformations and hence, in cases where these variations are significant, our method is not expected to perform well unless a sufficient number of instances are available in each configuration. Incorporating such invariances into our method cannot be done at a patch level and requires future work.

In Section~\ref{sec:method} we specified a number of parameter values used by our method. Some of the values do not play a critical role, such as the number of candidate patches sampled or the termination criterion of this search where higher values do not achieve significant improvement. Some scale values do not show dependance on the data as they are defined relative to the object size, such as the patch size, RANSAC diameter, and occluder's search distance. The value that may benefit further adaptation is the patch correlation tolerance $\varepsilon$ which may be depend on the level of noise or, perhaps in extreme cases, on the object's shape. Let us note that we produced all the results reported in the paper using the \emph{same} set of specified values. 

\section*{Acknowledgments}

\noindent This work was funded by the European Research Council, ERC Starting Grant 337383 ``Fast-Filtering". 
{\small
\bibliographystyle{ieee}
\bibliography{counting.bib}
}

\end{document}